\documentclass[10.5pt]{article}
\usepackage{indentfirst}
\usepackage{graphicx}
\usepackage{algorithm}
\usepackage{algorithmic}
\usepackage{fullpage}
\usepackage{epsf}
\usepackage{mdwlist}
\usepackage{mathrsfs}
\usepackage{amssymb}
\usepackage{amsthm}
\usepackage{bm}
\usepackage{tabularx,booktabs,multirow,delarray,array}
\newcommand{\func}[1]{{\textsf{#1}}}
\newcommand{\mf}[1]{{\mbox{\emph{#1}}}}
\newtheorem{define}{Definition}

\title{Cascading A*: a Parallel Approach to Approximate Heuristic Search}
\date{}
\author{Yan Gu}
\begin{document}
    \maketitle
    \par
\setcounter{secnumdepth}{2}
\setcounter{tocdepth}{2}

\section{Introduction and Background}

The main focus of this work is about a new special designed algorithm, called Cascading A* (CA*), to work efficiently on approximate heuristic graph search problems, especially on shortest path searching. The single source shortest path problem is defined as:

\begin{define}
Given a undirected graph $G=(V,E)$, two vertices are adjacent when they are both incident to a common edge. A path in an undirected graph is a sequence of vertices $P = ( v_1, v_2, \ldots, v_n ) \in V^n$ such that $v_i$ is adjacent to $v_{i+1}$ for $1 \leq i < n$. Such a path $P$ is called a path of length n from $v_1$ to $v_n$.

Let $e_{i, j}$ be the edge incident to both $v_i$ and $v_j$. Given a real-valued weight function $f: E \rightarrow \mathbb{R}$, and an undirected graph $G$, the shortest path from $v$ to $v'$ is the path $P = ( v_1, v_2, \ldots, v_n )$ that over all possible $n$ minimizes the sum $\sum_{i =1}^{n-1} f(e_{i, i+1})$, $v_1=v$ and $v_n=v'$.
\end{define}

A* algorithm and its variations are the most well-used algorithms on this problem. A* algorithm was first described by Peter Hart, Nils Nilsson and Bertram Raphael in 1968\,\cite{AStar}, which is an extension of Edsger Dijkstra's 1959 algorithm\,\cite{Dijkstra}. A* uses a best-first search and finds a least-cost path from a given initial node to one goal node. As A* traverses the graph, it follows a path of the lowest expected total cost or distance, keeping a sorted priority queue of alternate path segments along the way. It uses a knowledge-plus-heuristic cost function of node $x$ (usually denoted $f(x)$) to determine the order in which the search visits nodes in the tree. The cost function is a sum of two functions:
\begin{itemize}
  \item the past path-cost function, which is the known distance from the starting node to the current node $x$ (usually denoted $g(x)$).
  \item a future path-cost function, which is an admissible ``heuristic estimate'' of the distance from $x$ to the goal (usually denoted $h(x)$).
\end{itemize}
The $h(x)$ part of the $f(x)$ function is usually an admissible heuristic. Nevertheless, a non-admissible heuristic is used for computing approximate solution.

However, it is also widely accepted that the parallelism for general version A* is very difficult: A* keeps an global heap of candidate solutions, which involves heavy data concurrency if directly paralleled. Moreover, usually a hash table (or another similar data structure) is needed to avoid duplicate states, which is not scalable, since the search space usually increase exponentially as the increasing of the problem size.

Alternative versions of A* are proposed to overcome these problems, which includes IDA*\,\cite{IDAStar}, SMA*\,\cite{SMA}, beam search, etc. The most adopted version is Iterative Deepening A* (IDA*), which uses iterative deepening to keep the memory usage lower than in A*.
While the standard iterative deepening depth-first search uses search depth as the cutoff for each iteration the IDA* uses the more informative $f(n) = g(n) + h(n)$ where $g(n)$ is the cost to travel from the root to node $n$ and $h(n)$ is the heuristic estimate of the cost to travel from $n$ to the solution. However, IDA* suffers from the following 2 problems which lead it unsuitable for parallel approximate heuristic search problem:
\begin{itemize}
  \item The heuristic $f(n)$ here must be admissible, because IDA* will eventually search the whole space with the radius of the distance between the initial and goal state.
  \item Like the common problems in parallelizing DFS, it is usually hard to provide balanced workloads to different processors.
\end{itemize}

In this paper, we proposed an new algorithm, called Cascading A* (CA*) to overcome the shortcomings of the previous algorithms. Fully details of the algorithm will be explained in Section 3, and we state the advantages of our new algorithm here, which is also the contribution of our work:

\begin{itemize}
  \item An approximate approach. Unlike standard A* and most of its variants, CA* is able to generate approximate solution by using non-admissible heuristic, and trade off between running time and solution quality.
  \item Parallelism friendly. CA* is specially designed for parallel approaching. It can reach nearly linear speedup for modern CPUs. (8 cores for our experiments.)
  \item Any-time solution. CA* is able to generate sub-optimal solutions in a relatively fast speed, and find better solutions when keeping running. This is important advantage, since for the most real-world systems, responding time for a solution is usually more important than the quality of the solution.
\end{itemize}

\section{Cascading A*}

The Cascading A* algorithm is a hybrid algorithm of A* algorithm and IDA* algorithm. The high level idea of CA* algorithm is as following:
\begin{itemize}
        \item In order to quickly generate approximate solution, an outer level A* algorithm using non-admissible heuristic is used in CA*.
        \item For the sake of generating sufficient number of threads or sub-tasks for many-core hardware, we introduce the concept of the ``envelope ball'', which is an ball containing all the states or vertices with the radius of $R$, which is defined by a threshold function treated as a parameter of CA*. After any of the states reaching the envelop ball, it forks an new thread to solve the subtask using a separate core.
      \end{itemize}

After presenting the high level idea of the CA* algorithm, we provide the pseudocode in Algorithm \ref{main}.
\begin{algorithm}[h!]
\caption{$\func{Cascading A*(G, IS, GS))}$}
\label{main}
    Input: ~~~~~~~Graph $G=(V,E)$, either explicit or inexplicit; Initial State \mf{IS}, and Goal State \mf{GS}\\
    Parameter: \,Threshold function $t(S)$\\
    Output: ~~~~\,Approximate shortest path $P$
    \vspace{.3em}
    \begin{algorithmic}[1]
    \STATE {OpenSet = \{start\}};
    \STATE {CloseSet = \{\}};
    \STATE {GlobalAnswer = $+\infty$};
    \WHILE {OpenSet $\ne \varnothing$}
        \STATE {current = OpenSet.bestElement()};
        \FORALL {$S\in$ current.neighbor}
            \IF {($S\notin$ CloseSet) $\lor$ ($S$ can be optimal)}
                \IF {f(S)}
                    \STATE {IDA*(S)};
                \ELSE
                    \STATE Openset.add($S$);
                \ENDIF
            \ENDIF
        \ENDFOR
        \RETURN {GlobalAnswer};
    \ENDWHILE
   \end{algorithmic}
\end{algorithm}

The input of CA* includes a graph, a initial state and a goal state. Meanwhile, a threshold function $f(S)$ need to be given to construct the envelope ball. Finally, if any solution, either approximate or accurate, is found, the algorithm will return the best GlobalAnswer, which can be found by any thread.

There are two properties of CA*, which guarantees the advantages of CA* stated before. We will further discuss these properties in this paragraph.

\begin{itemize}
  \item CA* algorithm will activate multiple independent threads, and they are mostly not interactive with each other. Unlike traditional sequential implementations, CA* is significantly more parallel friendly, because nearly no data concurrency is needed for the communications of different cores. On the other hand, since CA* is a two-phrase algorithm, it can handle a much larger size of graph comparing with traditional implementations: if A* or IDA* is able to find a path in $V^n$, CA* can provide a reasonably high quality solution with a path in $V^m$, where $m\approx 2n$.
  \item Nevertheless, different threads do communicate and cooperate with each other, using the minimum cost of concurrency. This process is done by the only share global variable: GlobalAnswer. Once any of the thread find a new shortest path, the other threads can use this new upper bound to further prune their search tasks. This is especially powerful at the beginning, and this is the reason that sometimes the parallel version can reach the acceleration ratio which is larger than the number of the cores.
\end{itemize}

\section{Experiment Problem: The $N^2-1$ Sliding Puzzle}

In this section, we introduce the $N^2-1$ sliding puzzle, as the experiment problem of CA*.

The most well-known version is the fifteen puzzle. It was invented by Noyes Chapman and created a puzzle craze in 1880. The 15-puzzle consists of a frame of numbered square tiles in random order with one tile missing. The puzzle also exists in other sizes, particularly the smaller 8-puzzle. If the size is $3\times3$ tiles, the puzzle is called the 8-puzzle, and if $4\times4$ tiles, the puzzle is called the 15-puzzle named, respectively, for the number of tiles and the number of spaces. The object of the puzzle is to place the tiles in order (see diagram) by making sliding moves that use the empty space. An example is given in the following figure.
\begin{center}
\begin{figure}[!h]
\centering
  \includegraphics[width=0.6\columnwidth]{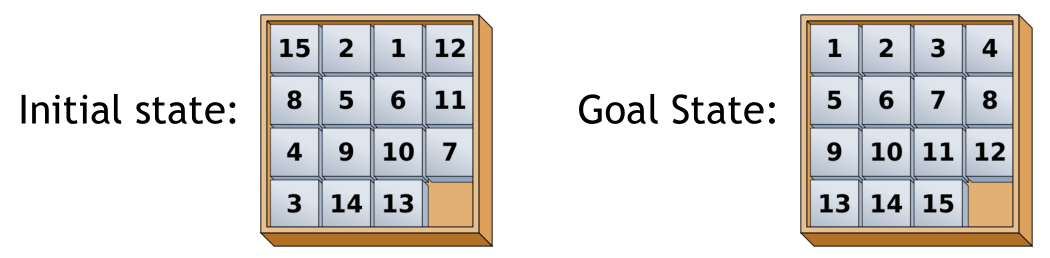}
  \caption{An example of 4-by-4 sliding puzzle.}
 \label{mesh}
\end{figure}
\end{center}

The $N^2-1$-puzzle is a classical problem for modelling algorithms involving heuristics. Commonly used heuristics for this problem is counting the sum of the Manhattan distances between each block and its position in the goal configuration.

However, the number of states of $5\times 5$ (24) puzzle is
$25!/2 = 7,755,605,021,665,492,992,000,000$. The search
space is too large to process exhausted search in near future. In 1996, Richard E. Korf and Larry A. Taylor\,\cite{c241} showed that 114 single-tile moves are needed. In 2000, Filip Karlemo and Patric Ostergard\,\cite{c242} showed that 210 single-tile moves are sufficient. The size of the graph is suitable for testing the performance of the CA* algorithm.

\section{Experiment and Results}

In order to test the performance of CA* algorithm, we choose to use $4\times 4$ (15) and $5\times 5$ (24) puzzles to test the performance.

We use the non-admissible heuristic $g+C\cdot h$ to get approximate solution, where $C$ is a constant between 1 to 2. Moreover, here, the threshold function we used is $t(S)=(h(S)<\theta_1)\,\lor\,((\mf{GlobalAnswer}-g(S)<\theta_2)\,\land\,(h(S)<\theta_3))$, where for most test cases, the parameters are $(\theta_1,\theta_2,\theta_3)=(30,40,15)$. Initially, CA* sets $\mf{GlobalAnswer}=+\infty$.

The machine we used for the experiment is ``oblivious.cs.cmu.edu'', which has 8 Intel i7 cores in 3.2 GHz. Parallel implementations were compiled with CilkPlus, which is included in g++.

10 puzzles, 5 for 15 puzzles and 5 for 24 puzzles, with different levels of difficulty, are used to measure the performance. To begin with, testing results from 15 puzzles are shown in Table 1.

\begin{table}[!h]
\begin{center}
\begin{tabular}{|rrrrrr|c|rrrr|}
\hline
           \multicolumn{ 6}{|c|}{Actual runtime of 15 puzzles (seconds)} & \multirow{2}{*}{Answer}  & \multicolumn{ 4}{c|}{Relative runtime of 15 puzzles} \\
\cline{1-6} \cline{8-11}
         \# case &     1 core &    2 cores &    4 cores &    8 cores &       IDA* &        &     1 core &    2 cores &    4 cores &    8 cores \\
\hline
         1 &       0.75 &       0.41 &       0.22 &       0.13 &        1.1 &         52 &          1.00 &       0.54 &       0.23 &       0.09 \\
\hline
         2 &        4.3 &        2.3 &        1.0 &        0.4 &        8.8 &         55 &          1.00 &        0.5 &       0.32 &       0.25 \\
\hline
         3 &        7.9 &        4.0 &        2.5 &        2.0 &       15.1 &         66 &          1.00 &       0.56 &       0.29 &       0.16 \\
\hline
         4 &       21.8 &       10.1 &        6.3 &        3.7 &       26.9 &         78 &          1.00 &       0.73 &       0.33 &       0.14 \\
\hline
         5 &       40.5 &       29.7 &       13.4 &        5.6 &       98.5 &         78 &          1.00 &       0.55 &       0.29 &       0.17 \\
\hline
\end{tabular}
\end{center}
\caption{Running time for all 5 test cases, with different number of cores in IDA*, and the running time for sequential version IDA*, relative runtime to sequential version Cascading A*, and the best answer to each test case.}
\end{table}

15 puzzle is smaller, so the IDA* algorithm can always give the solution. However, in all 5 cases, from the relatively easy case, to the hardest 15 puzzle, CA* gives the same answer as IDA*, but faster. The average relative running time for 5 test cases for CA* is 58\% comparing to the running time of IDA*. This is also shown in Figure 2.

For 24 puzzles, the size of the graph is too large to make exhausted search. Hence, in order to make the comparison to be fare, we manually set the cutoff answer, which means that whenever the algorithm reach this answer, the program terminates immediately, and the running time to get this answer is reported. This is because even if the code is the same, the code is not deterministic, and searches different states even if rerun the code in parallel.

For the running time here, we can find the CA* algorithm can reach nearly linear speedup on up to 8 cores, which is a big advantage for these kinds of searching algorithm. Moreover, by just comparing the running time and solution quality, the CA* is a very competitive heuristic graph searching algorithm.

\begin{center}
\begin{figure}[!h]
\centering
  \includegraphics[width=0.49\columnwidth]{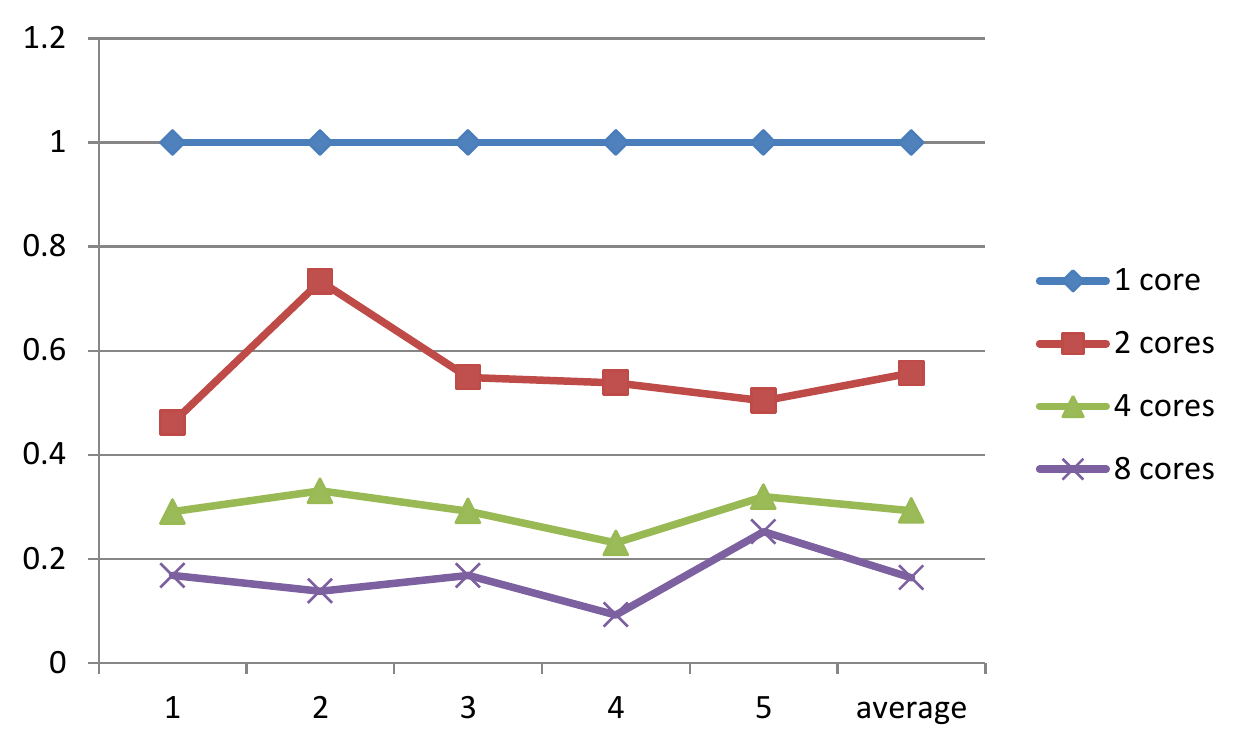}
  \includegraphics[width=0.49\columnwidth]{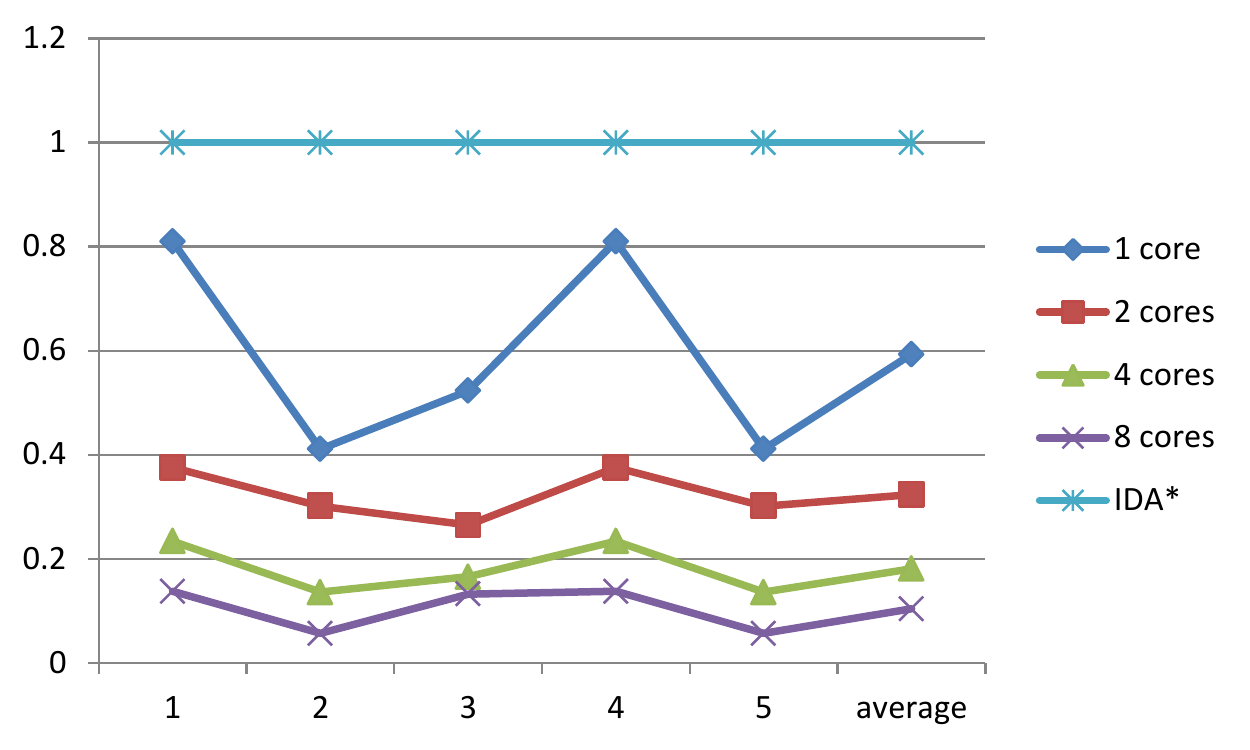}
  \caption{Relative running time to sequential version CA* (left), and to sequential IDA* (right). X-axis is 5 test cases and the average, and Y-axis is the relative time.}
 \label{mesh}
\end{figure}
\end{center}

\begin{table}[!h]
\begin{center}
\begin{tabular}{|rrrrr|c|rrrr|}
\hline
           \multicolumn{ 5}{|c|}{Actual runtime of 24 puzzles (seconds)} & \multirow{2}{*}{Answer}  & \multicolumn{ 4}{c|}{Relative runtime of 24 puzzles} \\
\cline{1-5} \cline{7-10}
         \# case &     1 core &    2 cores &    4 cores &    8 cores &         &     1 core &    2 cores &    4 cores &    8 cores \\
\hline
         1 &       25.4 &       11.5 &       10.9 &        7.3 &        122 &       1.00 &       0.45 &       0.43 &       0.29 \\
\hline
         2 &      241.3 &      122.7 &       46.7 &       26.5 &        264 &       1.00 &       0.51 &       0.19 &       0.11 \\
\hline
         3 &      122.7 &       81.3 &       22.7 &       10.0 &         60 &       1.00 &       0.66 &       0.18 &       0.08 \\
\hline
         4 &      115.3 &       52.7 &       32.0 &       17.3 &        106 &       1.00 &       0.46 &       0.28 &       0.15 \\
\hline
         5 &      152.7 &      104.0 &       54.7 &       22.0 &        145 &       1.00 &       0.68 &       0.36 &       0.14 \\
\hline
\end{tabular}
\end{center}
\caption{Running time for all 5 test cases, with different number of cores in IDA*, and the running time for sequential version IDA*, relative runtime to sequential version Cascading A*, and the cutoff answer to each test case.}
\end{table}

\begin{center}
\begin{figure}[!h]
\centering
  \includegraphics[width=0.49\columnwidth]{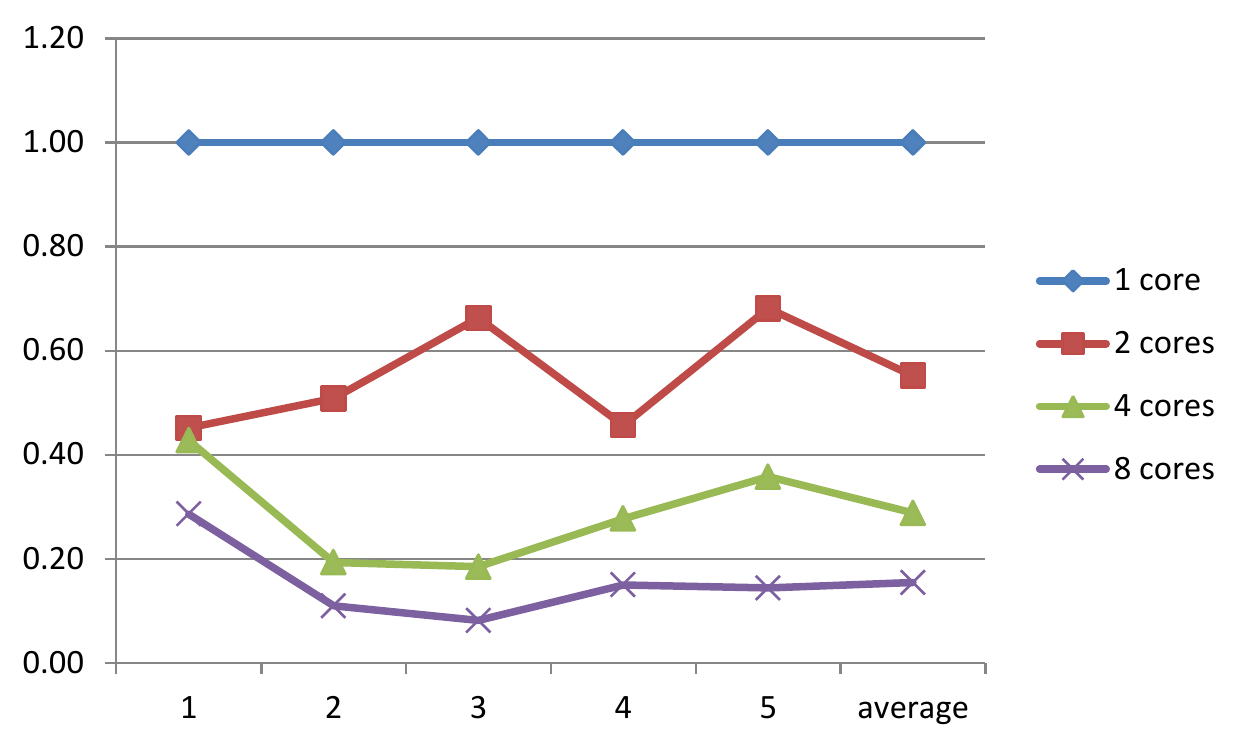}
  \caption{Relative running time of 24 puzzles to sequential version CA*. X-axis is 5 test cases and the average, and Y-axis is the relative time.}
 \label{mesh}
\end{figure}
\end{center}

\section{Conclusion}

In this paper, we proposed a new approximate heuristic search algorithm: Cascading A*, which is a two-phrase algorithm combining A* and IDA* by a new concept ``envelope ball''. The new algorithm CA* is efficient (comparing the running time to IDA* in 15 puzzle), able to generate approximate solution and any-time solution, and parallel friendly.

\end{document}